%% file: main.tex
\documentclass[10pt,twocolumn,letterpaper]{article}
\usepackage[pagenumbers]{cvpr}
\input{preamble}
\definecolor{cvprblue}{rgb}{0.21,0.49,0.74}
\usepackage[breaklinks,colorlinks,allcolors=cvprblue]{hyperref}
\hypersetup{pdftitle={DreamCount: Imagine Before You Count},
            pdfauthor={HaoYuan Ma, Shuai Deng, Taolue Meng, Li Zhang, Daizong Liu, Jianjun Li, Jie Gao, Zhiyuan Ma}}
\title{DreamCount: Imagine Before You Count}
\input{authors}
\begin{document}
\maketitle
\raggedbottom
\input{sec/0_abstract}

\input{sec/1_intro}
\input{sec/2_related}

\input{sec/3_benchmark}
\input{sec/4_method}
\input{sec/5_experiments}

\input{sec/6_conclusion}
\section*{LLM Usage Statement}
We used OpenAI's ChatGPT and Codex to refine sentence structure,
word choice, and paragraph transitions in the manuscript. The tools also helped us review the literature, check the text
for consistency, and prepare source files and figures.
We used GPT-Image2~\cite{openai2026gptimage2} to generate all synthetic
images in our benchmarks (Sec.~\ref{sec:benchmark}).
During annotation, GPT-5.6 Sol parsed whole--part hierarchies and
generated structured target descriptions. Grounding DINO provided
the initial bounding boxes. Human annotators reviewed
all released counts and bounding boxes. The authors reviewed and
verified the AI-assisted content and take responsibility for the
manuscript, data, and reported results.

{
    \small
    \bibliographystyle{ieeenat_fullname}
    \bibliography{main}
}

\end{document}

%% file: preamble.tex
\usepackage{multirow}
\usepackage{makecell}
\usepackage{booktabs}
\usepackage{colortbl}
\usepackage{enumitem}

\usepackage{algorithm}

\usepackage{algorithmic}

\definecolor{rowblue}{RGB}{246, 249, 254}

\definecolor{gaincyan}{HTML}{00BFFF}
\newcommand{\gain}[1]{{\fontsize{4.8pt}{4.8pt}\selectfont\color{gaincyan}\,$\downarrow$#1\%}}
\newcommand{\gainup}[1]{{\fontsize{4.8pt}{4.8pt}\selectfont\color{gaincyan}\,$\uparrow$#1\%}}
\definecolor{errorred}{HTML}{C33A32}
\newcommand{\errorincrease}[1]{{\fontsize{4.8pt}{4.8pt}\selectfont\color{errorred}\,$\uparrow$#1\%}}

\usepackage{microtype}

%% file: authors.tex
\author{
HaoYuan Ma$^{1,2}$ \quad Shuai Deng$^{3}$ \quad Taolue Meng$^{2}$ \quad Li Zhang$^{1,*}$\\
Daizong Liu$^{4}$ \quad Jianjun Li$^{2}$ \quad Jie Gao$^{1,*}$ \quad Zhiyuan Ma$^{2,*}$\\[0.4em]
{\small $^{1}$Soochow University \qquad $^{2}$Huazhong University of Science and Technology}\\
{\small $^{3}$South-Central Minzu University \qquad $^{4}$Wuhan University}\\
{\small $^{*}$Corresponding authors}
}

%% file: sec/0_abstract.tex
\begin{abstract}
Text-guided open-vocabulary object counting (TOOC) methods perform well in familiar domains but struggle when the requested counting unit changes, when targets change in appearance, or when they blend into the background. A target state derived from text may only partly capture the target content (the category and intended counting unit) or the target's appearance in a particular image.
We propose DreamCount, a model-agnostic framework for adapting a counting model's target state to the query and the image. Latent Dream generates target prototypes from Gaussian noise through conditional flow matching, using both text and image context. Lucid Guidance amplifies the difference between the text-conditioned and null-text velocities while keeping the image context fixed. Refine then adapts the prototypes to the observed target appearance using local visual evidence.
We also introduce \textup{\textsc{PolyCount}}, \textup{\textsc{WaterCount}}, and \textup{\textsc{CamoCount}} to evaluate granularity shifts, underwater appearance changes, and camouflage. Across five open-vocabulary hosts, DreamCount lowers mean absolute error (MAE) relative to the host baselines on the FSC-147 test split, on CARPK, and on all three new benchmarks. The largest reduction is 72.0\% for CountSE on CARPK. On REC-8K, it reduces test MAE by 5.0\% and 4.0\% for GroundingREC and CAD-GD, respectively. An interactive demo accompanies this preprint.
\end{abstract}

%% file: sec/1_intro.tex
\section{Introduction}
\label{sec:intro}

\begin{figure}[t]
\centering
\includegraphics[width=\linewidth]{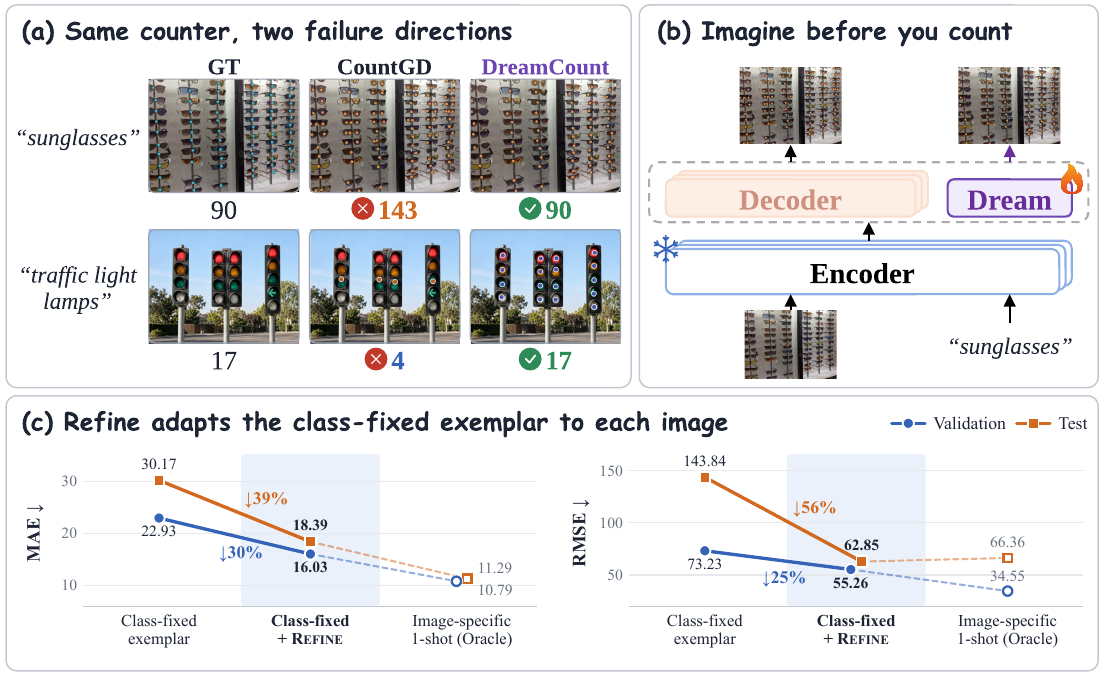}
\caption{\textbf{Open in vocabulary, closed in state.}
(a) CountGD overcounts sunglasses (143 vs.\ 90) and undercounts traffic lamps (4 vs.\ 17), illustrating whole--part ambiguity.
(b) \textsc{DreamCount} first generates a target state, then adapts it using localized visual evidence.
(c) Diagnostic evaluation of DAVE~\cite{DAVE} on FSC-147: refining the class-fixed exemplar reduces test MAE by 39.0\%.}
\label{fig:teaser}
\end{figure}

Text-guided open-vocabulary object counting (TOOC) aims to count objects from arbitrary categories specified in natural language, without requiring manually annotated exemplars at inference time~\cite{clipcount,vlcounter,countgd,countse}.
It supports scalable counting in applications such as inventory auditing~\cite{wei2019rpc,goldman2019precise} and wildlife monitoring~\cite{wildlife2016,wildlife2018}. Despite strong performance on standard benchmarks, existing methods struggle with changes in counting units or target appearance, as well as with objects that blend into their surroundings.
For example, CountGD~\cite{countgd} overcounts sunglasses (143 vs.\ 90) and undercounts individual traffic lamps (4 vs.\ 17) in \cref{fig:teaser}(a), illustrating ambiguity between whole objects and their parts.
These examples suggest that recognizing a category alone may not determine the intended counting unit.

To examine image-specific appearance, we evaluate DAVE~\cite{DAVE} with a class-fixed exemplar. We randomly select one exemplar per category and reuse it for all images of that category (\cref{fig:teaser}(c)).
In this controlled, training-free setting, test MAE is 11.29 with an image-specific 1-shot exemplar and 30.17 with a class-fixed exemplar.
Refinement reduces test MAE to 18.39, a 39.0\% reduction, suggesting that appearance refinement can help adapt the target state to the visual context of each image.
Together with the counting-unit examples above, this diagnostic motivates two complementary aspects of target-state construction: target content (the target category and intended counting unit) and image-specific appearance under conditions such as illumination changes, turbidity, or camouflage.

Existing approaches combine generated class prototypes with patch selection~\cite{zsc}, select soft exemplars using semantic similarity~\cite{countse}, or use pseudo-exemplars and external synthetic exemplars~\cite{countgdpp2026}.
These approaches motivate studying how target-state construction interacts with local evidence selection.

We propose \textsc{DreamCount}, a model-agnostic framework that first generates a state for the requested target content and then adapts it using local appearance evidence (\cref{fig:teaser}(b)).
Latent Dream generates this latent state from Gaussian noise through conditional flow matching~\cite{lipman2022flow}, conditioned on the text prompt and image context.
Lucid Guidance amplifies the contribution of text by comparing velocities computed with and without the query for the same image.
Refine then incorporates localized visual features to adapt the target state to the image.
At inference time, \textsc{DreamCount} requires only text prompts and test images.

To complement studies of group--individual and semantic granularity~\cite{bunchcount2026,kubricount}, we introduce three evaluation suites:
\textsc{PolyCount} for counting unit shifts across granularities,
\textsc{WaterCount} for underwater appearance changes, and
\textsc{CamoCount} for camouflage.
Across five open-vocabulary hosts and two referring-expression hosts, \textsc{DreamCount} lowers MAE in the reported evaluations, although gains vary by metric.

We make three contributions.
\begin{itemize}
    \item \textbf{Finding.} We highlight target content and image-specific appearance as complementary aspects of target-state construction. Controlled exemplar diagnostics show that appearance refinement reduces error relative to a class-fixed exemplar.
    \item \textbf{Framework.} We propose \textsc{DreamCount}, a model-agnostic framework that uses conditional flow matching to generate prototypes for the requested target and local image evidence to refine them.
    \item \textbf{Benchmarks.} We introduce \textsc{PolyCount}, \textsc{WaterCount}, and \textsc{CamoCount} to evaluate counting unit and appearance shifts across diverse counting models.
\end{itemize}

%% file: sec/2_related.tex
\section{Related Work}
\label{sec:related_work}

\subsection{Open-Vocabulary Object Counting}

\paragraph{Few-Shot Open-Vocabulary Object Counting.}
Few-shot counting specifies target categories through user-provided visual exemplars~\cite{fsc147}. Subsequent methods~\cite{bmnet,Countr,loca1,DAVE} improve exemplar--image correspondence through metric learning, cross-attention, and iterative verification. Text-guided counting aims to obtain rich target representations without manual exemplars at inference time.

\paragraph{Text-Guided Open-Vocabulary Object Counting.}
Text-guided open-vocabulary object counting (TOOC) uses text to specify targets.
CLIP-Count~\cite{clipcount} and VLCounter~\cite{vlcounter} adapt vision-language representations for counting, while CountGD~\cite{countgd} supports text and visual exemplars.
T2ICount~\cite{t2icount2025} predicts density maps using features extracted from a pretrained diffusion U-Net with \emph{one-step denoising}. It also uses hierarchical semantic correction and regional coherence supervision.
ZSC~\cite{zsc} uses a conditional variational autoencoder to generate features from class semantics. It averages those features to form a prototype, retrieves image patches by nearest-neighbor search, and uses predicted counting error for ranking.
CountSE~\cite{countse} selects multiscale soft exemplars using semantic similarity and clustering.
CountGD++~\cite{countgdpp2026} feeds high-confidence predicted boxes back as pseudo-exemplars and supports external natural or synthetic exemplars, including images generated from the input image and text.
\textsc{DreamCount} learns states in each host's target feature space using text and image context.
Latent Dream starts from Gaussian noise, integrates the learned velocity field in eight inference steps, and applies a learned residual update. Refine uses the resulting state to select local regions, then updates each matched slot with a weighted average of its prior and the corresponding regional feature.
Training combines the native counting loss with flow-matching supervision, coupling state generation and local adaptation across density- and detection-based hosts.

\subsection{Existing Counting Datasets and Benchmarks}
Counting benchmarks have expanded from crowd~\cite{MCNN} and vehicle~\cite{carpk} counting to exemplar-guided counting on FSC-147~\cite{fsc147} and FSCD-LVIS~\cite{FSCD-LVIS}, referring-expression counting on REC-8K~\cite{REC}, taxonomic counting on TPC-268~\cite{TPC-268}, and multi-task evaluation on UNICBench~\cite{Unicbench}.
BunchCount~\cite{bunchcount2026} directly examines group--individual ambiguity, including part--whole cases, using paired annotations and containment relations across 1,330 real images; its GICount model uses counting-unit prompts and relational supervision.
KubriCount~\cite{kubricount} evaluates identity, attribute, category, instance-type, and concept queries; HieraCount combines text and exemplar prompts to localize the intended instances.
These works establish query-dependent granularity as a common challenge.
We contribute host-compatible target-state generation and local adaptation, evaluated on \textsc{PolyCount} using paired whole--part queries and on the separate camouflage and underwater suites, \textsc{CamoCount} and \textsc{WaterCount}.

%% file: sec/3_benchmark.tex
\begin{table*}[t]
\centering
\caption{\textbf{Comparison with existing counting benchmarks.}
Our benchmarks cover paired whole--part queries, underwater optical distortion, and camouflage. Related multi-granularity benchmarks are discussed in Sec.~\ref{sec:related_work}. WaterCount is summarized in terms of its image--query pairs.}
\label{tab:dataset_comparison}
\small
\setlength{\tabcolsep}{6.0pt}
\begin{tabular}{lcccccc}
\toprule
\textbf{Dataset} & \textbf{Venue} & \textbf{Images} & \textbf{Instances} & \textbf{Mean Count} & \textbf{Resolution} ($W\!\times\!H$) & \textbf{Categories} \\
\midrule
\multicolumn{7}{l}{\emph{Class-Specific Counting Benchmarks}} \\
ShanghaiTech Part A~\cite{MCNN} & CVPR'16 & 482 & 241,677 & 501.40 & 868$\times$589 & 1 \\
CARPK~\cite{carpk} & ICCV'17 & 1,448 & 89,777 & 62.00 & 1280$\times$720 & 1 \\
NWPU-Crowd~\cite{nwpu} & TPAMI'21 & 5,109 & 2,133,375 & 417.57 & 3209$\times$2191 & 1 \\
IOCfish5K~\cite{iocfish5k} & CVPR'23 & 5,637 & 659,024 & 116.91 & 1920$\times$1080 & 1 \\
\midrule
\multicolumn{7}{l}{\emph{Open-World and Language-Guided Counting Benchmarks}} \\
FSC-147~\cite{fsc147} & CVPR'21 & 6,135 & 344,150 & 56.10 & 523$\times$384 & 147 \\
FSCD-LVIS~\cite{FSCD-LVIS} & ECCV'22 & 6,196 & 402,945 & 65.03 & 586$\times$479 & 377 \\
REC-8K~\cite{REC} & CVPR'24 & 8,011 & 286,621 & 35.78 & 640$\times$480 & 80 (8k expr.) \\
TPC-268~\cite{TPC-268} & CVPR'26 & 10,000 & 678,050 & 67.81 & 1130$\times$959 & 268 \\
\midrule
\multicolumn{7}{l}{\emph{Proposed Target-State Mismatch Benchmarks}} \\
\textbf{PolyCount (whole)} & This work & 691 & 2,597 & 3.76 & 384$\times$384 & 37 \\
\textbf{PolyCount (part)} & This work & 691 & 13,508 & 19.55 & 384$\times$384 & 37 \\
\textbf{CamoCount} & This work & 538 & 4,692 & 8.72 & 384--942$\times$384--684 & 81 \\
\textbf{WaterCount} & This work & \multicolumn{5}{l}{1,007 image--query pairs across 10 underwater categories} \\
\bottomrule
\end{tabular}
\end{table*}

\begin{figure}[t]
\centering
\includegraphics[width=\columnwidth]{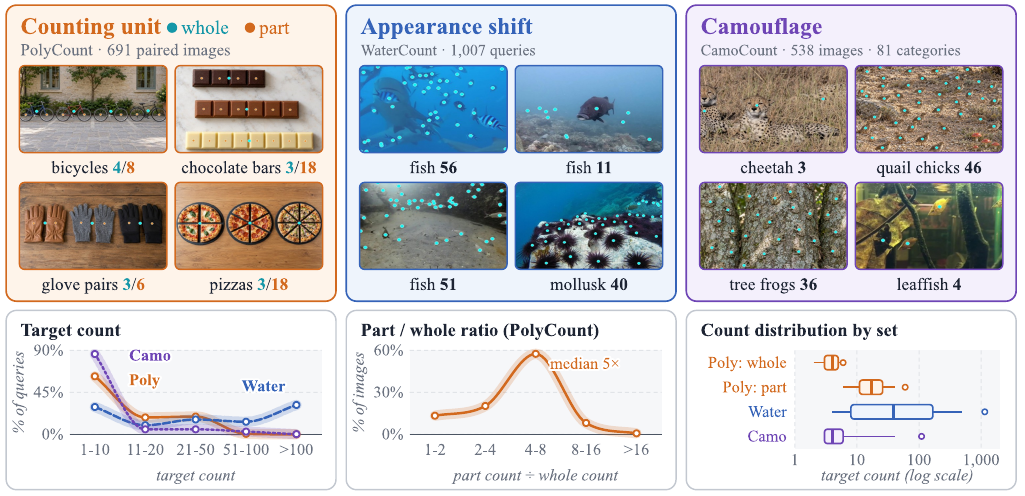}
\caption{\textbf{Examples from the three proposed benchmarks.}
\textsc{PolyCount} (left) pairs targets at two granularities within the same scene (\eg, whole bikes \vs wheels, chocolate bars \vs pieces);
\textsc{WaterCount} (middle) covers underwater optical degradation, including turbidity, backscatter, and color attenuation;
\textsc{CamoCount} (right) contains targets that blend into their surroundings.}
\label{fig:benchmarks}
\end{figure}
\section{Benchmark Construction}
\label{sec:benchmark}

\paragraph{Overview.}
We introduce three complementary benchmarks to systematically evaluate target state mismatch (\cref{fig:benchmarks}):
\textsc{PolyCount} for counting unit shifts,
\textsc{WaterCount} for underwater appearance changes,
and \textsc{CamoCount} for camouflage.
\Cref{tab:dataset_comparison} summarizes their statistics alongside those of existing benchmarks.

\paragraph{Data Construction.}
We collected photographs from verified public sources and datasets. We used GPT-Image2~\cite{openai2026gptimage2} to generate all synthetic images in our three benchmarks, targeting conditions underrepresented in the photographs.

\textbf{PolyCount.}
PolyCount contains 691 images and 1,382 textual queries arranged in pairs spanning 37 whole--part category hierarchies. The dataset has 370 real images from the web (53.5\%) and 321 synthetic scenes (46.5\%).
Each image is annotated with a pair of counting units at two granularities (\eg, ``battery packs'' \vs ``batteries''). This pairing isolates counting unit shifts while keeping the visual scene fixed (\cref{fig:benchmarks}, left).

\textbf{WaterCount.}
WaterCount comprises 1,007 image--query pairs across 10 underwater categories, with photographs from IOCfish5K~\cite{iocfish5k} and UIIS10K~\cite{UIIS10k} as well as scenes generated with GPT-Image2. It evaluates counting under underwater appearance shifts caused by turbidity, backscatter, and color attenuation (\cref{fig:benchmarks}, middle).

\textbf{CamoCount.}
CamoCount contains 538 multi-instance camouflage scenes with 4,692 instances across 81 categories. These include 464 real images (86.2\%) from CAMO~\cite{CAMO}, CAMO-FS~\cite{CAMO-FS}, and UCIS4K~\cite{UCIS4K}, together with 74 scenes generated using GPT-Image2 (13.8\%) for underrepresented camouflaged species. The benchmark evaluates target disambiguation when object boundaries blend into the background (\cref{fig:benchmarks}, right).

\paragraph{Annotation and Quality Control.}
To initialize high-precision annotations, we build an automated annotation agent that combines GPT-5.6 Sol~\cite{openai2026gpt56sol} with Grounding DINO-Base~\cite{GroundingDINO}. GPT-5.6 Sol parses whole--part hierarchies and generates structured target descriptions, and Grounding DINO produces initial instance bounding boxes.
A team of five annotators then manually reviews every sample to correct missing detections, false positives, and inconsistent counting units.
Multiple annotators resolve ambiguous cases by consensus, so all released counts and bounding boxes undergo human verification.

\paragraph{Evaluation Protocol.}
In the main experiments, we train all base methods and \textsc{DreamCount} exclusively on FSC-147~\cite{fsc147}.
Without target-domain fine-tuning, we evaluate on the complete query sets: 691 images / 1,382 queries on \textsc{PolyCount}, 1,007 image--query pairs on \textsc{WaterCount}, and 538 queries on \textsc{CamoCount}.
We also define category-disjoint training, validation, and test splits for reference evaluations. Their sizes are 484/105/102 images for \textsc{PolyCount}, 336/168/34 images for \textsc{CamoCount}, and 934/55/18 image--query pairs for \textsc{WaterCount}.

%% file: sec/4_method.tex
\section{Method}
\label{sec:method}

\begin{figure}[t]
\centering
\includegraphics[width=\columnwidth]{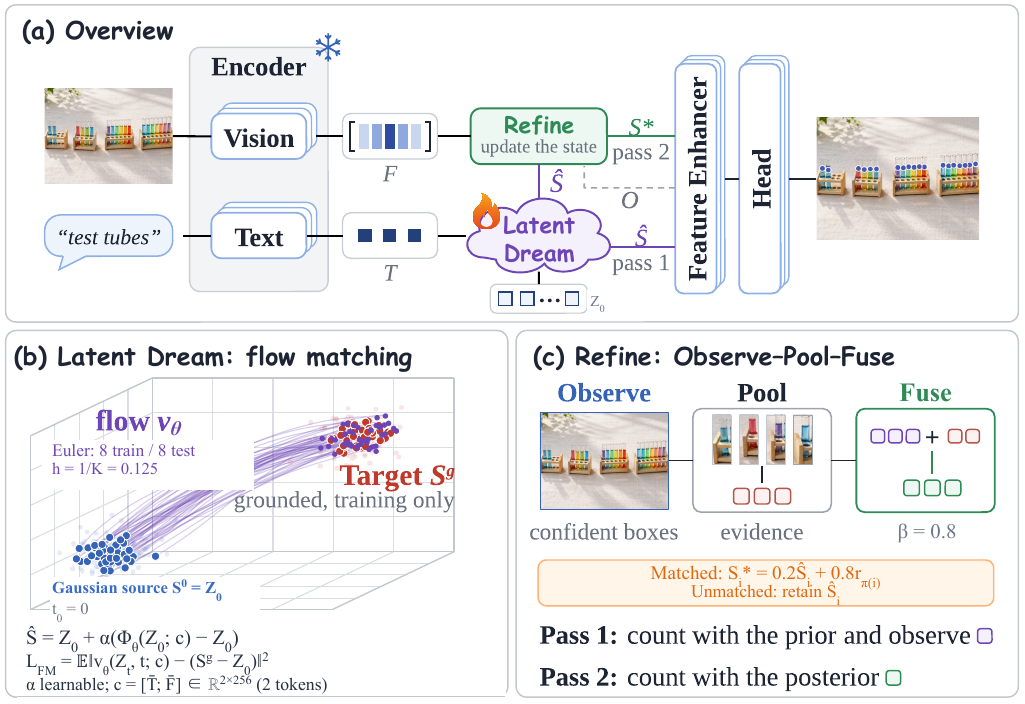}
\caption{\textbf{Architectural overview of the \textsc{DreamCount} framework.}
(a) Given an image $I$ and a text query $u$, \textsc{DreamCount} generates and then refines a target state.
(b) Starting from Gaussian noise, Latent Dream integrates the guided velocity field from $t=0$ to $1$ using text and image context. A learned residual update yields the prior $\hat{S}$.
(c) The training-free Refine step blends matched prior slots with local visual evidence ($\beta=0.8$), retains unmatched priors, and passes the adapted state $S^*$ to the Feature Enhancer.}
\label{fig:framework}
\label{fig:architecture}
\end{figure}

\textsc{DreamCount} is a model-agnostic imagine-then-refine pipeline (\cref{fig:framework}(a)).
Latent Dream generates prototypes for the requested target content using text and image context. Refine adapts these prototypes to local appearance using visual evidence.
The resulting state $S^*$ replaces the generic text embedding $T$ in the host's Feature Enhancer for final prediction.

\subsection{Preliminaries: Text-Guided Object Counting}
\label{sec:preliminaries}

Given an image $I$ and a text query $u$, the visual and text encoders produce the respective features
\begin{equation}
    F = E_v(I) \in \mathbb{R}^{H \times W \times 256}, \qquad T = E_t(u) \in \mathbb{R}^{L \times 256}.
\end{equation}
A host counting model consists of a cross-modal Feature Enhancer $\mathcal{E}_{\mathrm{feat}}$ and a Prediction Head $\mathcal{H}$.
The model predicts object counts conditioned on image features $F$ and text embedding $T$:
\begin{equation}
    F_{\mathrm{enh}} = \mathcal{E}_{\mathrm{feat}}(F, T), \qquad Y = \mathcal{H}(F_{\mathrm{enh}}),
\end{equation}
The host output $Y$ consists of scored boxes for CountGD~\cite{countgd} and CountSE~\cite{countse}, points for RT-Counter~\cite{RT-Counter}, or a density map for CounTX~\cite{CounTX} and CLIP-Count~\cite{clipcount}.

\textsc{DreamCount} reconstructs $S$ directly in latent space:

\begin{equation}
\begin{aligned}
    \hat{S} &= \operatorname{LatentDream}(Z_0; c), \\
    S^* &= \operatorname{Refine}(F, \hat{S}), \\
    Y &= \mathcal{H}\big(\mathcal{E}_{\mathrm{feat}}(F, S^*)\big).
\end{aligned}
\end{equation}

Here, $Z_0$ is Gaussian noise, $c$ is the global conditioning signal, and $\hat{S}$ is the imagined prior. We denote the host's original text-conditioned target state by $S_{\mathrm{def}}$.
Refine adapts $\hat{S}$ using image features $F$, combining the generated target state with localized appearance evidence to form $S^*$, which directly replaces the generic text embedding $T$ in the Feature Enhancer.

\subsection{Latent Dream: Target-State Imagination}
\label{sec:latent_dream}

To mitigate target content mismatch, Latent Dream uses flow matching to learn a distribution of target states in latent space, conditioned on text and image context.

\paragraph{Grounded Prototype Construction.}
During training, we extract 256-dimensional target features from ground-truth instance boxes using RoIAlign \cite{loca1}.
We construct a grounded prototype set by randomly sampling one target feature for each of the three prototype slots:
\begin{equation}
    S^g = R = [r^g_1, r^g_2, r^g_3] \in \mathbb{R}^{3 \times 256}.
\end{equation}
These grounded prototypes supervise the velocity field during training and remain strictly unavailable at inference time.

\paragraph{Conditional Flow Matching.}
We sample a Gaussian source state $Z_0 \sim \mathcal{N}(0, I)$ and define the linear interpolation path:
\begin{equation}
    Z_t = (1 - t) Z_0 + t R, \qquad t \sim \mathcal{U}(0, 1).
\end{equation}
The constant target velocity along this path is $u_t = R - Z_0$.

Let $\bar{T}$ be the mean of the valid text embeddings and $\bar{F}$ the mean of the visual features at valid spatial positions.
We stack the two pooled features as separate context tokens,
\begin{equation}
    c = [\bar{T}; \bar{F}] \in \mathbb{R}^{2\times256}.
\end{equation}

The velocity network $v_\theta(Z_t, t; c)$ uses self-attention across prototype slots and cross-attention over context $c$ to predict the velocity that transports Gaussian noise toward grounded target prototypes.

\paragraph{Lucid Guidance.}
Dropping all conditions removes image context from the null branch.
During training, Lucid Guidance replaces only the text condition with a learnable null-text token $\varnothing$ with probability $0.1$, while retaining the global image feature $\bar{F}$.

At inference, we combine conditioned and null-text velocities:
\begin{equation}
    v_{\mathrm{CFG}} = v_u + w\,(v_c - v_u),
\end{equation}
where $w$ is the guidance scale, and
\begin{equation}
\begin{aligned}
    v_c &= v_\theta(Z_t, t; \bar{T}, \bar{F}), \\
    v_u &= v_\theta(Z_t, t; \varnothing, \bar{F}).
\end{aligned}
\end{equation}
Here, $v_c$ is the text-conditioned velocity and $v_u$ is the null-text velocity for the same image.
With $w=1.5$, guidance amplifies the velocity difference $(v_c-v_u)$ induced by the text query. We assess its effect on counting granularity empirically.

\paragraph{Prototype Sampling.}
At inference, we sample $S^0=Z_0 \sim \mathcal{N}(0,I)$ for each image, initialize $Z^{(0)}=Z_0$ at $t_0=0$, and integrate to $t_K=1$:
\begin{equation}
    Z^{(j+1)} = Z^{(j)} + h\,v_{\mathrm{CFG}}(Z^{(j)}, t_j), \quad h=\frac{1}{K},
\end{equation}
where $t_j=jh$ for $j=0,\ldots,K-1$. At test time, we use $K=8$ ($h=0.125$).

Let $\Phi_\theta$ denote this guided Euler mapping, with terminal state:
\begin{equation}
    \Phi_\theta(S^0;c) = Z^{(K)}.
\end{equation}
We form the imagined prior using a learnable residual coefficient $\alpha$:
\begin{equation}
    \hat{S} = S^0 + \alpha\big(\Phi_\theta(S^0;c)-S^0\big).
    \label{eq:dream_residual}
\end{equation}

The imagined prior state $\hat{S} \in \mathbb{R}^{3 \times 256}$ is then passed to Refine to guide feature-level adaptation.

\subsection{Refine: Image-Grounded Feature Adaptation and Fusion}
\label{sec:refine}

\begin{figure}[t]
\centering
\includegraphics[width=\columnwidth]{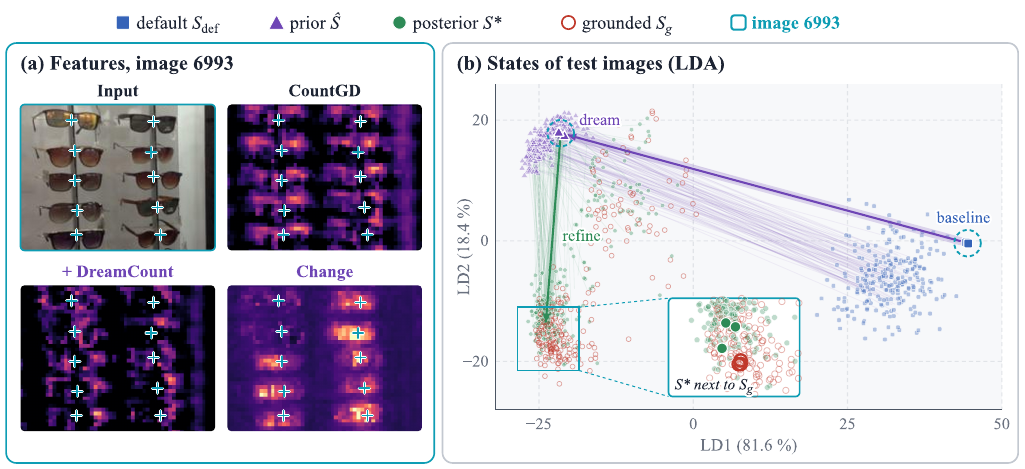}
\caption{\textbf{Feature activation shift and latent state trajectory.}
(a) In the FSC-147 sunglasses image, CountGD's activations concentrate on individual lenses. \textsc{DreamCount} shifts the activations toward whole pairs of sunglasses (\emph{Change}).
(b) A linear discriminant analysis (\textsc{LDA}) projection shows the host's default states $S_{\mathrm{def}}$, imagined priors $\hat{S}$, refined states $S^*$, and oracle grounded targets $S^g$ across test images. The host uses $S_{\mathrm{def}}$ as its default state, and Dream starts from the Gaussian source $S^0$.}
\label{fig:features}
\end{figure}

Although Latent Dream uses global image conditioning and image-specific training targets, the generated prototypes may still lack local instance-level appearance details.
\Cref{fig:features}(a) shows the shift in activations toward whole objects, and \cref{fig:features}(b) traces the latent states toward the grounded targets. Refine uses $\hat{S}$ and $F$ to extract local appearance evidence and construct the adapted state $S^*$, which it passes to the Feature Enhancer (\cref{fig:framework}(c)).

\paragraph{Observe.}
For detection and point hosts, a preliminary pass conditioned on $\hat{S}$ supplies boxes or centers for candidate windows. We retain up to 50 candidates with confidence above $0.5$. For density hosts (\eg, CounTX and CLIP-Count), we score each valid position $p$ by $\max_i\cos(\hat{S}_i,F_p)$ and center candidate windows at local maxima of this score.

\paragraph{Pool.}
For each candidate region $b_q$, we extract an instance-level visual feature vector from the feature map $F$:
\begin{equation}
    r_q = \operatorname{RoIAlign}_{1 \times 1}(F, b_q) \in \mathbb{R}^{256}.
\end{equation}
We compute cosine similarities between each candidate feature vector and each imagined prototype slot:
\begin{equation}
    a_{iq} = \cos(\hat{S}_i, r_q) = \frac{\hat{S}_i^\top r_q}{\|\hat{S}_i\| \|r_q\|}, \qquad i \in \{1, 2, 3\}.
\end{equation}

\paragraph{Fuse.}
We retain candidate pairs satisfying $a_{iq} \ge 0.5$ and perform greedy bipartite matching in descending order of similarity.
Each prototype slot and candidate feature participates in at most one assignment $\pi(i)$.
We fuse each matched prototype with its visual evidence using a fixed weight $\beta=0.8$:
\begin{equation}
S_i^* = \begin{cases}
    (1-\beta)\hat{S}_i+\beta r_{\pi(i)}, & \text{if matched}, \\
    \hat{S}_i, & \text{otherwise}.
\end{cases}
\end{equation}
For a matched slot, the prior has weight $20\%$ and the regional feature has weight $80\%$. Each unmatched slot retains its prior.

\paragraph{Predict.}
The adapted target state $S^*$ directly replaces the ambiguous text embedding $T$ and, together with the visual feature map $F$, is fed into the Feature Enhancer $\mathcal{E}_{\mathrm{feat}}$:
\begin{equation}
    F_{\mathrm{enh}} = \mathcal{E}_{\mathrm{feat}}(F, S^*).
\end{equation}
The Prediction Head $\mathcal{H}$ then produces the final counting output from $F_{\mathrm{enh}}$.
CountGD and CountSE count retained detection boxes using host-specific counting thresholds. For CountGD, the count from its 900 decoder queries is
\begin{equation}
    \hat{N} = \sum_{q=1}^{900} \mathbb{I}\left[ \max_a \sigma(\ell_{qa}^*) > \tau_{\mathrm{count}} \right].
\end{equation}
Point hosts count retained foreground points. Density hosts integrate the final density map $D^*$ to obtain $\hat{N} = \iint D^*(x, y) \, dx dy$.
Here, the similarity threshold $0.5$ filters candidate matches for prototype adaptation, whereas the host-specific enumeration threshold $\tau_{\mathrm{count}}$ controls the precision--recall trade-off of the final count.

\subsection{Optimization Objectives}
\label{sec:optimization}

We freeze the visual and text encoders and jointly optimize Latent Dream, including $\alpha$, and each host's remaining trainable components using its original training recipe. Our pipeline replaces CountSE's SE component.

\paragraph{Flow Matching Loss.}
Let $\tilde{T}$ denote the text condition after dropout. The flow loss is
\begin{equation}
\mathcal{L}_{\mathrm{FM}} = \mathbb{E}_{Z_0,R,t,\tilde{T}} \bigl[\bigl\|v_\theta(Z_t,t;\tilde{T},\bar{F})-u_t\bigr\|_2^2\bigr].
\end{equation}

\paragraph{Overall Objective.}
The total training objective combines the host's native loss with the flow matching loss:
\begin{equation}
\mathcal{L}_{\mathrm{train}} = \mathcal{L}_{\mathrm{count}} + 0.05\,\mathcal{L}_{\mathrm{FM}},
\end{equation}
where $\mathcal{L}_{\mathrm{count}}$ is the host's standard counting loss.
Ground-truth target features $R$ supervise the velocity field during training and are never accessed during inference.

%% file: sec/5_experiments.tex
\begin{table*}[!t]
\centering
\caption{\textbf{Comprehensive evaluation across five diverse datasets.}
The top panel reports in-domain counting on \textsc{FSC-147} and zero-shot transfer to \textsc{CARPK}.
The bottom panel reports zero-shot transfer under changes in counting unit (\textsc{PolyCount}), camouflage (\textsc{CamoCount}), and underwater appearance shifts (\textsc{WaterCount}).
All five host methods are trained solely on FSC-147 and evaluated without downstream fine-tuning.
For CountSE, ``+\textsc{DreamCount}'' replaces the SE component.
We compute relative changes in error from the displayed scores and report GSA changes in percentage points. Arrows indicate the direction of change. Cyan and red indicate improvements and declines, respectively. Bold marks the better result for each pair.}
\label{tab:main_benchmarks}
\vspace{2pt}
\small
\setlength{\tabcolsep}{5pt}
\begin{tabular}{l l c c c c c c}
\toprule
\multirow{2}{*}{\textbf{Method}} & \multirow{2}{*}{\textbf{Configuration}} & \multicolumn{2}{c}{\textbf{FSC-147 Val}} & \multicolumn{2}{c}{\textbf{FSC-147 Test}} & \multicolumn{2}{c}{\textbf{CARPK Zero-Shot}} \\
\cmidrule(lr){3-4} \cmidrule(lr){5-6} \cmidrule(lr){7-8}
& & \textbf{MAE} $\downarrow$ & \textbf{RMSE} $\downarrow$ & \textbf{MAE} $\downarrow$ & \textbf{RMSE} $\downarrow$ & \textbf{MAE} $\downarrow$ & \textbf{RMSE} $\downarrow$ \\
\midrule
  & Baseline & 10.87 & 48.74 & 11.26 & 103.42 & 10.30 & 13.49 \\
  \rowcolor{rowblue}
\multirow{-2}{*}{CountSE~\cite{countse}} & + \textsc{DreamCount} & \textbf{9.94}\gain{8.6} & \textbf{47.28}\gain{3.0} & \textbf{10.78}\gain{4.3} & \textbf{98.76}\gain{4.5} & \textbf{2.88}\gain{72.0} & \textbf{4.01} \\
\midrule
  & Baseline & \textbf{19.69} & 64.52 & 18.01 & \textbf{104.65} & 16.71 & \textbf{20.41} \\
  \rowcolor{rowblue}
\multirow{-2}{*}{CLIP-Count~\cite{clipcount}} & + \textsc{DreamCount} & 20.85\errorincrease{5.9} & \textbf{64.31}\gain{0.3} & \textbf{17.53}\gain{2.7} & 107.29\errorincrease{2.5} & \textbf{15.86}\gain{5.1} & 20.85 \\
\midrule
  & Baseline & 17.36 & 62.46 & 18.54 & 108.49 & 12.81 & 15.81 \\
  \rowcolor{rowblue}
\multirow{-2}{*}{CounTX~\cite{CounTX}} & + \textsc{DreamCount} & \textbf{16.24}\gain{6.5} & \textbf{59.64}\gain{4.5} & \textbf{17.69}\gain{4.6} & \textbf{105.31}\gain{2.9} & \textbf{11.74}\gain{8.4} & \textbf{14.66} \\
\midrule
  & Baseline & 12.56 & 51.25 & 13.30 & 104.63 & 4.81 & 7.61 \\
  \rowcolor{rowblue}
\multirow{-2}{*}{RT-Counter~\cite{RT-Counter}} & + \textsc{DreamCount} & \textbf{11.85}\gain{5.7} & \textbf{50.72}\gain{1.0} & \textbf{13.00}\gain{2.3} & \textbf{103.19}\gain{1.4} & \textbf{4.52}\gain{6.0} & \textbf{6.84} \\
\midrule
  & Baseline & 12.89 & 51.08 & 12.63 & 107.47 & 4.19 & 5.64 \\
  \rowcolor{rowblue}
\multirow{-2}{*}{CountGD~\cite{countgd}} & + \textsc{DreamCount} & \textbf{9.42}\gain{26.9} & \textbf{47.88}\gain{6.3} & \textbf{10.13}\gain{19.8} & \textbf{103.58}\gain{3.6} & \textbf{3.79}\gain{9.5} & \textbf{5.26} \\
\bottomrule
\end{tabular}

\vspace{5pt}
\small
\setlength{\tabcolsep}{1.7pt}
\begin{tabular}{l l c c c c c c c c c}
\toprule
\multirow{2}{*}{\textbf{Method}} & \multirow{2}{*}{\textbf{Configuration}} & \multicolumn{2}{c}{\textbf{\textsc{PolyCount} Whole}} & \multicolumn{2}{c}{\textbf{\textsc{PolyCount} Part}} & \textbf{GSA} & \multicolumn{2}{c}{\textbf{\textsc{CamoCount}}} & \multicolumn{2}{c}{\textbf{\textsc{WaterCount}}} \\
\cmidrule(lr){3-4} \cmidrule(lr){5-6} \cmidrule(lr){8-9} \cmidrule(lr){10-11}
& & \textbf{MAE} $\downarrow$ & \textbf{RMSE} $\downarrow$ & \textbf{MAE} $\downarrow$ & \textbf{RMSE} $\downarrow$ & \textbf{(\%)} $\uparrow$ & \textbf{MAE} $\downarrow$ & \textbf{RMSE} $\downarrow$ & \textbf{MAE} $\downarrow$ & \textbf{RMSE} $\downarrow$ \\
\midrule
  & Baseline & 34.20 & 46.85 & 12.15 & 22.40 & 18.4 & 10.36 & 28.62 & 51.61 & 111.46 \\
  \rowcolor{rowblue}
\multirow{-2}{*}{CountSE~\cite{countse}} & + \textsc{DreamCount} & \textbf{12.40}\gain{63.7} & \textbf{23.10}\gain{50.7} & \textbf{11.80}\gain{2.9} & \textbf{21.90}\gain{2.2} & \textbf{81.5}\gainup{63.1} & \textbf{5.95}\gain{42.6} & \textbf{14.15}\gain{50.6} & \textbf{49.83}\gain{3.4} & \textbf{102.67}\gain{7.9} \\
\midrule
  & Baseline & 42.15 & 55.40 & 19.80 & 31.25 & 15.2 & 52.66 & 97.80 & 82.42 & 137.35 \\
  \rowcolor{rowblue}
\multirow{-2}{*}{CLIP-Count~\cite{clipcount}} & + \textsc{DreamCount} & \textbf{18.30}\gain{56.6} & \textbf{29.75}\gain{46.3} & \textbf{18.90}\gain{4.5} & \textbf{29.50}\gain{5.6} & \textbf{76.8}\gainup{61.6} & \textbf{49.76}\gain{5.5} & \textbf{83.29}\gain{14.8} & \textbf{74.72}\gain{9.3} & \textbf{130.99}\gain{4.6} \\
\midrule
  & Baseline & 38.60 & 51.20 & 18.20 & 29.40 & 22.8 & 59.22 & 89.33 & 94.94 & 144.38 \\
  \rowcolor{rowblue}
\multirow{-2}{*}{CounTX~\cite{CounTX}} & + \textsc{DreamCount} & \textbf{17.10}\gain{55.7} & \textbf{28.15}\gain{45.0} & \textbf{17.40}\gain{4.4} & \textbf{28.10}\gain{4.4} & \textbf{80.2}\gainup{57.4} & \textbf{33.36}\gain{43.7} & \textbf{51.02}\gain{42.9} & \textbf{92.64}\gain{2.4} & \textbf{126.01}\gain{12.7} \\
\midrule
  & Baseline & 31.80 & 44.50 & 13.90 & 24.15 & 24.6 & 27.48 & 123.27 & 71.05 & 130.96 \\
  \rowcolor{rowblue}
\multirow{-2}{*}{RT-Counter~\cite{RT-Counter}} & + \textsc{DreamCount} & \textbf{13.90}\gain{56.3} & \textbf{24.60}\gain{44.7} & \textbf{13.20}\gain{5.0} & \textbf{23.40}\gain{3.1} & \textbf{83.4}\gainup{58.8} & \textbf{12.66}\gain{53.9} & \textbf{53.16}\gain{56.9} & \textbf{70.40}\gain{0.9} & \textbf{129.23}\gain{1.3} \\
\midrule
  & Baseline & 28.45 & 39.60 & 11.85 & 20.30 & 28.4 & 9.97 & 31.88 & 55.63 & 119.48 \\
  \rowcolor{rowblue}
\multirow{-2}{*}{CountGD~\cite{countgd}} & + \textsc{DreamCount} & \textbf{11.85}\gain{58.3} & \textbf{21.40}\gain{46.0} & \textbf{10.90}\gain{8.0} & \textbf{19.80}\gain{2.5} & \textbf{88.6}\gainup{60.2} & \textbf{8.25}\gain{17.3} & \textbf{24.86}\gain{22.0} & \textbf{53.08}\gain{4.6} & \textbf{115.25}\gain{3.5} \\
\bottomrule
\end{tabular}
\end{table*}

\section{Experiments}
\label{sec:experiments}

\subsection{Experimental Setup}
\label{sec:exp_setup}

\paragraph{Data, Base Methods, and Metrics.}
We evaluate \textsc{DreamCount} across five open-vocabulary methods: CountSE~\cite{countse}, CLIP-Count~\cite{clipcount}, CounTX~\cite{CounTX}, RT-Counter~\cite{RT-Counter}, and CountGD~\cite{countgd}.
Plug-in variants are trained solely on FSC-147~\cite{fsc147} ($3{,}659/1{,}286/1{,}190$ train/val/test images) without downstream annotations.
We evaluate the complete query sets of \textsc{PolyCount}, \textsc{WaterCount}, and \textsc{CamoCount}, and also report zero-shot transfer to CARPK.
We report mean absolute error (MAE) and root mean squared error (RMSE).
For \textsc{PolyCount}, let $\hat y_q$ denote the prediction for query $q$, $y_q$ its ground-truth count, and $y_{\bar q}$ the count for the other unit in the same image. Granularity Switch Accuracy (GSA) over the query set $\mathcal Q$ is
\begin{equation}
\mathrm{GSA}=\frac{100}{|\mathcal Q|}\sum_{q\in\mathcal Q}\mathbf{1}\!\left[|\hat y_q-y_q|<|\hat y_q-y_{\bar q}|\right].
\end{equation}
Each whole or part query receives equal weight. A query succeeds when the prediction is closer to its own ground-truth count than to the count for the other unit. Ties count as failures, and we impose no absolute-error tolerance. We report a single run.

\paragraph{Training and Test Protocol.}
Each host retains its native losses and trainable modules outside the frozen encoders.
We reproduce CountSE from its official open-source implementation and train both the baseline and our SE replacement in the same environment. Our baseline MAE is $11.26$ on FSC-147 Test and $10.30$ on CARPK. CountSE reports $7.84$ and $2.79$, respectively, under its published protocol~\cite{countse}. We compare our replacement with the baseline reproduced here.
For CountGD, we train for 30 epochs with AdamW ($\text{lr}=10^{-4}$, weight decay $10^{-4}$, $\lambda_{\mathrm{FM}}=0.05$, FP32), use one image per optimizer step, and select the checkpoint by validation MAE. Training and inference both use eight Euler steps ($h=0.125$, $t_0=0$).
For Refine, we use the lowest-error setting from the FSC-147 Test sweep, $\beta=0.8$ and $\tau_{\mathrm{sim}}=0.5$. CountGD and CountSE count detection boxes using host-specific thresholds. CountGD inference uses an 800 px short side (long side capped at 1,333 px), $w=1.5$, and $\tau_{\mathrm{count}}=0.35$ on an NVIDIA Tesla V100.

\begin{figure*}[!t]
\centering
\includegraphics[width=0.98\textwidth]{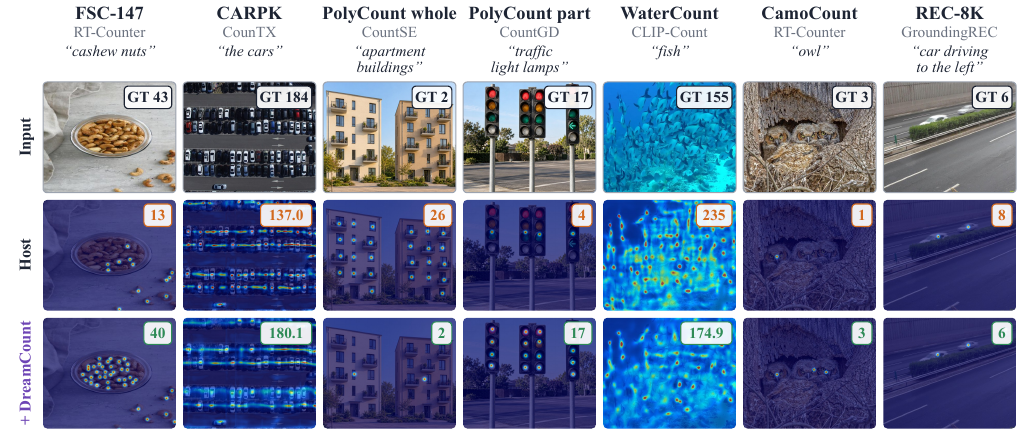}
\vspace{-8pt}
\caption{\textbf{Qualitative comparison across seven evaluation settings.}
Rows show the input, host prediction, and prediction with \textsc{DreamCount}; labels give counts for selected examples illustrating unit selection, adverse appearance, and localization.}
\label{fig:qualitative}
\vspace{-6pt}
\end{figure*}

\begin{figure*}[!t]
\centering
\includegraphics[width=0.70\textwidth]{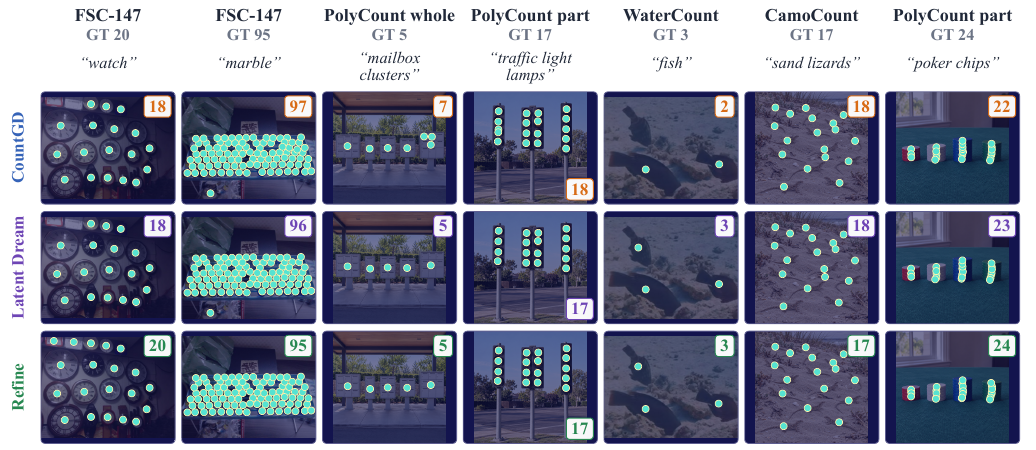}
\vspace{-6pt}
\caption{\textbf{CountGD predictions in a stage-wise diagnostic.}
Rows show the baseline, Latent Dream, and one round of local refinement at $\tau_{\mathrm{count}}=0.38$. This diagnostic replaces matched prototypes with local features. The main CountGD evaluation uses weighted fusion ($\beta=0.8$) and $\tau_{\mathrm{count}}=0.35$.}
\label{fig:qualitative_stages}
\vspace{2pt}
\end{figure*}

\subsection{Main Benchmark Results}
\label{sec:main_results}

\paragraph{FSC-147 and CARPK Benchmark Evaluations.}
\Cref{tab:main_benchmarks} shows lower MAE for all five hosts on FSC-147 Test and in zero-shot transfer to CARPK. \Cref{fig:qualitative} shows individual predictions.
CountGD's FSC-147 Test MAE decreases from $12.63$ to $10.13$ ($19.8\%$). On CARPK, CountSE's MAE decreases from $10.30$ to $2.88$ ($72.0\%$), while reductions for the other hosts range from $5.1\%$ to $9.5\%$. These gains are relative to our reproduced baselines; the CountSE protocol discrepancy precludes a direct ranking against its published numbers.

\paragraph{PolyCount Multi-Granularity Disambiguation.}
Zero-shot transfer from FSC-147 to \textsc{PolyCount} Whole tests difficult granularity shifts without target-domain adaptation.
In \cref{tab:main_benchmarks}, the baselines have whole \textsc{MAE} of $28.45$--$42.15$ and \textsc{GSA} of $15.2\%$--$28.4\%$.
\textsc{DreamCount} reduces whole-query \textsc{MAE} by $55.7\%$--$63.7\%$ across hosts. For example, whole-query MAE falls from $28.45$ to $11.85$ for CountGD and from $34.20$ to $12.40$ for CountSE.
Part \textsc{MAE} improves by $2.9\%$--$8.0\%$ across hosts, suggesting that the larger whole-level gains reflect improved unit selection while retaining sensitivity to finer targets.
Absolute error remains high. Predicting zero on every whole query gives an MAE of $3.76$, the mean whole count, compared with $11.85$--$18.30$ for the augmented hosts. Identifying persistent overcounting, outliers, or query mapping errors requires individual predictions.

\paragraph{Adverse Domain Generalization: Camouflage and Underwater Optics.}
On \textsc{CamoCount}, both missed targets and spurious detections can affect the count (\cref{tab:main_benchmarks}). For nonnegative predictions, undercounting contributes at most the mean target count ($8.72$) to MAE. For CounTX (baseline MAE $59.22$), the lower bound on the share due to overcounting is about $85.3\%$. Identifying the visual cause requires individual predictions. \textsc{DreamCount} reduces RT-Counter's MAE and RMSE by $53.9\%$ and $56.9\%$, respectively, and CountSE's by $42.6\%$ and $50.6\%$.
On \textsc{WaterCount}, all five hosts improve in MAE and RMSE, including a $12.7\%$ RMSE reduction for CounTX ($144.38\to126.01$). MAE reductions range from $0.9\%$ to $9.3\%$, smaller than on \textsc{PolyCount} Whole, indicating more limited gains under underwater appearance shifts.

\begin{table}[t]
\caption{\textbf{Zero-shot referring expression counting on REC-8K.}
Bold indicates the better result within each model and split.}
\label{tab:rec8k_main}
\small
\centering
\setlength{\tabcolsep}{2.0pt}
\renewcommand{\arraystretch}{0.85}
\resizebox{\columnwidth}{!}{%
\begin{tabular}{@{}l c r r r r r@{}}
\toprule
\textbf{Configuration} & \textbf{Split} & \textbf{MAE} $\downarrow$ & \textbf{RMSE} $\downarrow$ & \textbf{Prec} $\uparrow$ & \textbf{Rec} $\uparrow$ & \textbf{F1} $\uparrow$ \\
\midrule
\multicolumn{7}{l}{GroundingREC~\cite{groundingrec2024}} \\
Baseline & Val & 5.97 & 17.12 & 65.58 & \textbf{72.06} & \textbf{68.67} \\
\rowcolor{rowblue}
+ \textsc{DreamCount} & Val & \textbf{5.85} & \textbf{16.94} & \textbf{71.63} & 63.39 & 67.26 \\
Baseline & Test & 6.00 & 19.44 & \textbf{72.67} & 66.47 & 69.43 \\
\rowcolor{rowblue}
+ \textsc{DreamCount} & Test & \textbf{5.70} & \textbf{18.79} & 67.40 & \textbf{73.16} & \textbf{70.16} \\
\midrule
\multicolumn{7}{l}{CAD-GD~\cite{CAD-GD}} \\
Baseline & Val & 6.32 & 15.58 & 64.65 & 68.54 & 66.54 \\
\rowcolor{rowblue}
+ \textsc{DreamCount} & Val & \textbf{6.27} & \textbf{15.22} & \textbf{65.31} & \textbf{70.74} & \textbf{67.92} \\
Baseline & Test & 6.26 & 19.23 & \textbf{67.70} & 70.86 & 69.24 \\
\rowcolor{rowblue}
+ \textsc{DreamCount} & Test & \textbf{6.01} & \textbf{17.97} & 67.62 & \textbf{71.90} & \textbf{69.69} \\
\bottomrule
\end{tabular}%
}
\end{table}

\paragraph{Referring Expression Counting on REC-8K.}
We use each host's original evaluation protocol (\cref{tab:rec8k_main}). On GroundingREC, \textsc{DreamCount} reduces test MAE from $6.00$ to $5.70$ and raises recall from $66.5\%$ to $73.2\%$. On CAD-GD, test MAE falls from $6.26$ to $6.01$, recall and F1 rise, and precision changes little ($67.7\%$ to $67.6\%$).

\subsection{Ablation Studies and Analysis}
\label{sec:ablations}
We evaluate the components on CountGD using FSC-147 and \textsc{PolyCount}.

\begin{table}[t]
\caption{\textbf{\textsc{DreamCount} component ablation.}
Evaluation on CountGD with a retrained baseline.}
\label{tab:ablations}
\small
\centering
\setlength{\tabcolsep}{2.0pt}
\renewcommand{\arraystretch}{0.82}
\resizebox{\columnwidth}{!}{%
\begin{tabular}{@{}l r r r r@{}}
\toprule
\multirow{2}{*}{\textbf{Configuration}} & \multicolumn{2}{c}{\textbf{FSC-147 Test}} & \multicolumn{2}{c}{\textbf{PolyCount}} \\
\cmidrule(lr){2-3} \cmidrule(lr){4-5}
& \textbf{MAE} $\downarrow$ & \textbf{RMSE} $\downarrow$ & \textbf{Whole MAE} $\downarrow$ & \textbf{Part MAE} $\downarrow$ \\
\midrule
\multicolumn{5}{l}{\emph{Component Dissection}} \\
Baseline (same-budget) & 12.63 & 107.47 & 28.45 & 11.85 \\
Dream only & 10.19 & 111.65 & 13.24 & 11.73 \\
Refine only & 10.85 & \textbf{103.22} & 15.36 & 11.49 \\
\rowcolor{rowblue}
\textbf{Dream $\to$ Refine (Full)} & \textbf{10.13} & 103.58 & \textbf{11.85} & \textbf{10.90} \\
\bottomrule
\end{tabular}%
}
\end{table}

\begin{table}[t]
\caption{\textbf{Target-state generator variants.}
CountGD on FSC-147 and \textsc{PolyCount} Whole. $\dagger$: non-flow baseline.}
\label{tab:generator_variants}
\small
\centering
\setlength{\tabcolsep}{2.2pt}
\renewcommand{\arraystretch}{0.82}
\resizebox{\columnwidth}{!}{%
\begin{tabular}{@{}l r r r r@{}}
\toprule
\multirow{2}{*}{\textbf{Configuration / Variant}} & \multicolumn{2}{c}{\textbf{FSC-147 Test}} & \multicolumn{2}{c}{\textbf{PolyCount Whole}} \\
\cmidrule(lr){2-3} \cmidrule(lr){4-5}
& \textbf{MAE} $\downarrow$ & \textbf{RMSE} $\downarrow$ & \textbf{MAE} $\downarrow$ & \textbf{RMSE} $\downarrow$ \\
\midrule
Deterministic MLP$^\dagger$ & 10.78 & 107.22 & 12.34 & \textbf{16.90} \\
Multi-Head ($M{=}4$)$^\dagger$ & 10.53 & 108.91 & \textbf{11.35} & 16.97 \\
\rowcolor{rowblue}
\textbf{Ours (\textsc{DreamCount})} & \textbf{10.13} & \textbf{103.58} & 11.85 & 21.40 \\
\bottomrule
\end{tabular}%
}
\end{table}

\paragraph{Qualitative Stage Progression.}
\Cref{fig:qualitative_stages} follows CountGD predictions through Dream and a replacement-based refinement step. \Cref{tab:ablations} evaluates the components with the weighted fusion used in the main experiments.

\paragraph{Component Contributions and Generator Variants.}
Dream alone reduces whole MAE from $28.45$ to $13.24$. Refine alone gives $15.36$ and the lowest FSC-147 RMSE ($103.22$). Together they yield the lowest MAE on FSC-147 ($10.13$) and on whole and part queries ($11.85$ and $10.90$). Adding Refine to Dream lowers MAE by $10.5\%$ on wholes and $7.1\%$ on parts, compared with $0.6\%$ on FSC-147. It also reduces FSC-147 RMSE from $111.65$ to $103.58$, reversing the increase with Dream alone.
Among the generators in \cref{tab:generator_variants}, flow matching gives the lowest FSC-147 MAE and RMSE ($10.13/103.58$). On \textsc{PolyCount} Whole, the multi-head generator achieves lower MAE ($11.35$ vs.~$11.85$), and the MLP achieves lower RMSE ($16.90$ vs.~$21.40$).

%% file: sec/6_conclusion.tex
\section{Conclusion}
\label{sec:conclusion}
Target-state mismatch remains a challenge for open-vocabulary counting under granularity shifts and adverse optical conditions. \textsc{DreamCount} mitigates mismatches in target content and local appearance without requiring manual exemplars at inference. It generates prototypes through flow matching conditioned on text and image context, then refines them using local visual evidence. Across five open-vocabulary hosts, it lowers MAE relative to the corresponding baselines in the main evaluation. Ablations on CountGD suggest that imagination and refinement are particularly complementary under granularity shifts. Despite these gains, absolute errors on \textsc{PolyCount} whole queries remain substantial (MAE 11.85--18.30), and improvements vary across metrics. These findings motivate further analysis of individual predictions and more reliable adaptation to the requested counting unit.